\documentclass{article}

\usepackage{arxiv}
\PassOptionsToPackage{numbers, compress}{natbib}

\usepackage{amsmath,amsfonts,bm}

\def\eqref#1{equation~\ref{#1}}

\def\1{\bm{1}}

\DeclareMathAlphabet{\mathsfit}{\encodingdefault}{\sfdefault}{m}{sl}
\SetMathAlphabet{\mathsfit}{bold}{\encodingdefault}{\sfdefault}{bx}{n}

\usepackage[utf8]{inputenc} 
\usepackage[T1]{fontenc}    
\usepackage{hyperref}       
\usepackage{url}            
\usepackage{booktabs}       
\usepackage{amsfonts}       
\usepackage{nicefrac}       
\usepackage{microtype}      
\usepackage{graphicx}
\usepackage{natbib}
\usepackage{doi}
\usepackage{xcolor}         
\usepackage{tikz}
\usepackage{threeparttable}
\usepackage{adjustbox}

\usepackage{amsthm}  

\usepackage{array}
\usepackage{makecell}
\usepackage{multirow}
\usepackage{graphicx}
\usepackage{subcaption}
\usepackage{comment}
\usepackage{wrapfig}	
\usepackage{tabularx}
\usepackage{caption}
\usepackage{booktabs}
\usepackage[tableposition=top]{caption}
\usepackage{float}
\floatstyle{plaintop}
\restylefloat{table}
\usepackage{makecell}
\usepackage{titlesec}
\titlespacing\section{-2pt}{-2pt}{-2pt}
\titlespacing\subsection{-3pt}{-3pt}{-3pt}
\titlespacing\subsubsection{-3pt}{-3pt}{-3pt}
\titlespacing\paragraph{1pt}{1pt}{1pt}
\usepackage[capitalize,noabbrev]
{cleveref}

\usepackage{xspace}

\makeatletter
\DeclareRobustCommand\onedot{\futurelet\@let@token\@onedot}
\def\@onedot{\ifx\@let@token.\else.\null\fi\xspace}

\makeatother

\usepackage{enumitem}
\usepackage[titletoc,toc,page]{appendix}

\title{Video Relationship Detection Using Mixture of Experts}

\author{Ala Shaabana\\
	Department of Systems Design Engineering\\
	University of Waterloo\\
	Waterloo, ON  N2L 3G1 \\
	\texttt{ala.shaabana@uwaterloo.ca} \\
	\And
	Zahra Gharaee \\
	Department of Systems Design Engineering\\
	University of Waterloo\\
	Waterloo, ON  N2L 3G1 \\
	\texttt{zahra.gharaee@uwaterloo.ca} \\
 	\And
	Paul Fieguth \\
	Senior Member, IEEE\\
	Department of Systems Design Engineering\\
	University of Waterloo\\
	Waterloo, ON  N2L 3G1 \\
	\texttt{paul.fieguth@uwaterloo.ca} \\
}

\date{}

\renewcommand{\headeright}{}
\renewcommand{\undertitle}{}
\renewcommand{\shorttitle}{MoE-VRD}

\hypersetup{
pdftitle={SoftCD arxiv},
pdfsubject={q-bio.NC, q-bio.QM},
pdfauthor={Shayan Shirahmad},
pdfkeywords={Motion forecasting, Causality, Domain Adaptation},
}

\begin{document}
\maketitle

\begin{abstract}
Machine comprehension of visual information from images and videos by neural networks faces two primary challenges. Firstly, there exists a computational and inference gap in connecting vision and language, making it difficult to accurately determine which object a given agent acts on and represent it through language. Secondly, classifiers trained by a single, monolithic neural network often lack stability and generalization. To overcome these challenges, we introduce \textbf{MoE-VRD}, a novel approach to visual relationship detection utilizing a mixture of experts. MoE-VRD identifies language triplets in the form of $<\text{\em subject}, \text{\em predicate}, \text{\em object}>$ tuples to extract relationships from visual processing. Leveraging recent advancements in visual relationship detection, MoE-VRD addresses the requirement for action recognition in establishing relationships between subjects (acting) and objects (being acted upon). In contrast to single monolithic networks, MoE-VRD employs multiple small models as experts, whose outputs are aggregated. Each expert in MoE-VRD specializes in visual relationship learning and object tagging. By utilizing a sparsely-gated mixture of experts, MoE-VRD enables conditional computation and significantly enhances neural network capacity without increasing computational complexity. Our experimental results demonstrate that the conditional computation capabilities and scalability of the mixture-of-experts approach lead to superior performance in visual relationship detection compared to state-of-the-art methods.

\end{abstract}


\section{Introduction}
\label{sec:intro}
In the last decade, there has been a surge in research on the machine comprehension of visual information from images and video sequences.  In particular, the application of large neural networks has allowed problems to be tackled such as video object segmentation \cite{caelles2017one,perazzi2017learning,tokmakov2017learning,pont20172017}, object recognition and classification \cite{yan2006discriminative, zhang2007real, han2016seq, kang2016object, karpathy2014large}, and action recognition \cite{feichtenhofer2017spatiotemporal, carreira2017quo,8620199,8626113,8121994}.

This unprecedented progress in the comprehension of visual information, however, suffers from the computational and inference gap between vision and language \cite{shang2021video} to accurately determine which object a given agent acts on and how it might represent it by language.

We can begin by drawing inspiration from the two-streams hypothesis of the brain and how it processes visual information, such that the brain distinguishes between a ventral stream (the ``what'' pathway) and a dorsal stream (the ``where'' or ``how'' pathway) \cite{goodale1992separate}. In parallel with this distinction, natural languages contain two classes of verbs describing actions:  manner verbs, describing {\em how} an action is performed by expressing cause, such as waving arms (implying cheering) or nodding head (for assent); as opposed to result verbs, that describe the {\em result} of an action by expressing their effect, such as move, heat, clean, enter etc.~\cite{warglien2012event}. 

Computationally, then, several approaches to visual information processing focus on the manner in which an action is performed (how) \cite{arunnehru2018human,gharaee2017first,muhammad2021human,gharaee2020hierarchical}, versus the result of the action, whether an object moves (where) or changes in appearance \cite{gharaee2017online,giorgi2021learning}. 

When an action results in nearby changes, the visual information processing problem consists of detecting the three inter-related entities of subject, predicate (action), and the object(s) involved; that is, to recognize language triplets in the form of a $<\!\text{\em subject}, \text{\em predicate}, \text{\em object}\!>$ tuple.  To detect the relationship between a subject (acting) and the object(s) (acted upon), the action must be recognized. Some recent approaches to visual relationship detection have focused on static images \cite{tsai2019video}, however static relationship detection clearly has limitations in understanding temporal constraints inherent in video sequences, which offer significant richness regarding relationships \cite{tsai2019video, shang2017video, shang2021video}. Therefore, there has been a significant research emphasis on detecting visual relationships in video sequences \cite{shang2017video, qian2019video, tsai2019video, liu2020beyond, su2020video, XiaoSYTC2020, shang2021video}.

The main challenges associated with this problem stem from the very large datasets, high ambiguity, and huge amount of background clutter. Moreover, the objects involved may only  barely be recognizable due to pose, motion blur, occlusion and lighting. On the other hand, large variance in predicate representations \cite{shang2017video} also makes it difficult to learn latent patterns, thus it is essential to consider visual and spatial features, and language ambiguity with synonyms. There is also a combinatorial effect, in that the number of unique tuple classes can be exceptionally large (the product of the vocabulary of subjects, objects, and predicates).

In this article, we propose an approach to video visual relationship detection (VidVRD), implemented by a multi-expert framework, where each expert is trained using the same model, and where the outputs of all experts are gated based on a separate neural network. Our performance results show that our novel architecture substantially outperforms known state-of-the-art methods. The contributions of this work are thus two-fold:
\begin{enumerate}
	\item We construct a novel multi-expert detection framework.
	\item We capture recent developments in video visual relationship detection as experts in our proposed multi-expert architecture.  Our proposed approach is not tuned to a particular choice of expert, and other choices of expert should be equally valid and applicable.
\end{enumerate}
The rest of this paper is organized as follows: Section~\ref{related_works} further develops the VidVRD problem and overviews past work, Section~\ref{sect:arch} describes the encapsulation of an existing VidVRD approach into an expert \cite{shang2021video}, Section~\ref{experiments} describes the experiments and results which are discussed in Section~\ref{discuss}.

\section{Related work}
\label{related_works}

Cascading failures caused by low-level misclassifications make a monolithic solution, such as a single very large neural network, not an ideal strategy to resolve large visual relationship detection problems.  To address such limitations, earlier approaches proposed solutions such as dividing a given problem based on pre-processing and post-processing heuristics \cite{shang2017video, shang2021video,tsai2019video,gharaee2021bayesian}, to attempt to correctly classify predicates. 

The capacity of a neural network to learn is limited by its degrees of freedom (number of parameters) and further limited by the available data. When datasets are, in fact, large enough, then increasing the number of parameters can lead to significant improvements in performance.  However for a typical deep learning model, where the entire network is activated on each training sample, the computational cost is roughly quadratic in the number of parameters, as both the model size and the number of training samples increase together \cite{davis2013low,bengio2015conditional,shazeer2017}.  This phenomenon is particularly and uniquely exacerbated in visual information processing since the input layer is very large. The statistical relationships of pixels and objects detected in images and videos are subtle, and the networks are often expected to perform multiple tasks like object segmentation and action recognition, all of which lead to significant increases in network size.

Limitations in computing power will eventually
fall short of meeting the training demand, and the networks trained in this fashion tend to be brittle and sensitive to
slight changes in the data distribution \cite{recht2018cifar}
and task specification \cite{kirkpatrick2017overcoming}.  That is, current systems are better characterized as {\em narrow experts} rather than as {\em resilient generalists} \cite{radford2019language}.

With the above context in mind, we make the following observation about the VidVRD task: A training approach based on a single, monolithic network is not sufficient to have stable and generalizable classifiers, at least in certain problem contexts \cite{recht2018cifar,radford2019language}. Therefore, applying divide-and-conquer by developing multiple small models and aggregating their outputs could be a promising solution \cite{dong2020survey,yin2017recognition} to create more compact and/or resilient networks. As a consequence, the mixture-of-experts architecture suggests a strategy to achieve the larger network capacity needed to solve large numbers of sub-problems, which is typical of the VidVRD domain.

\begin{figure*}[tp]
	\centering
	\includegraphics[scale=.5]{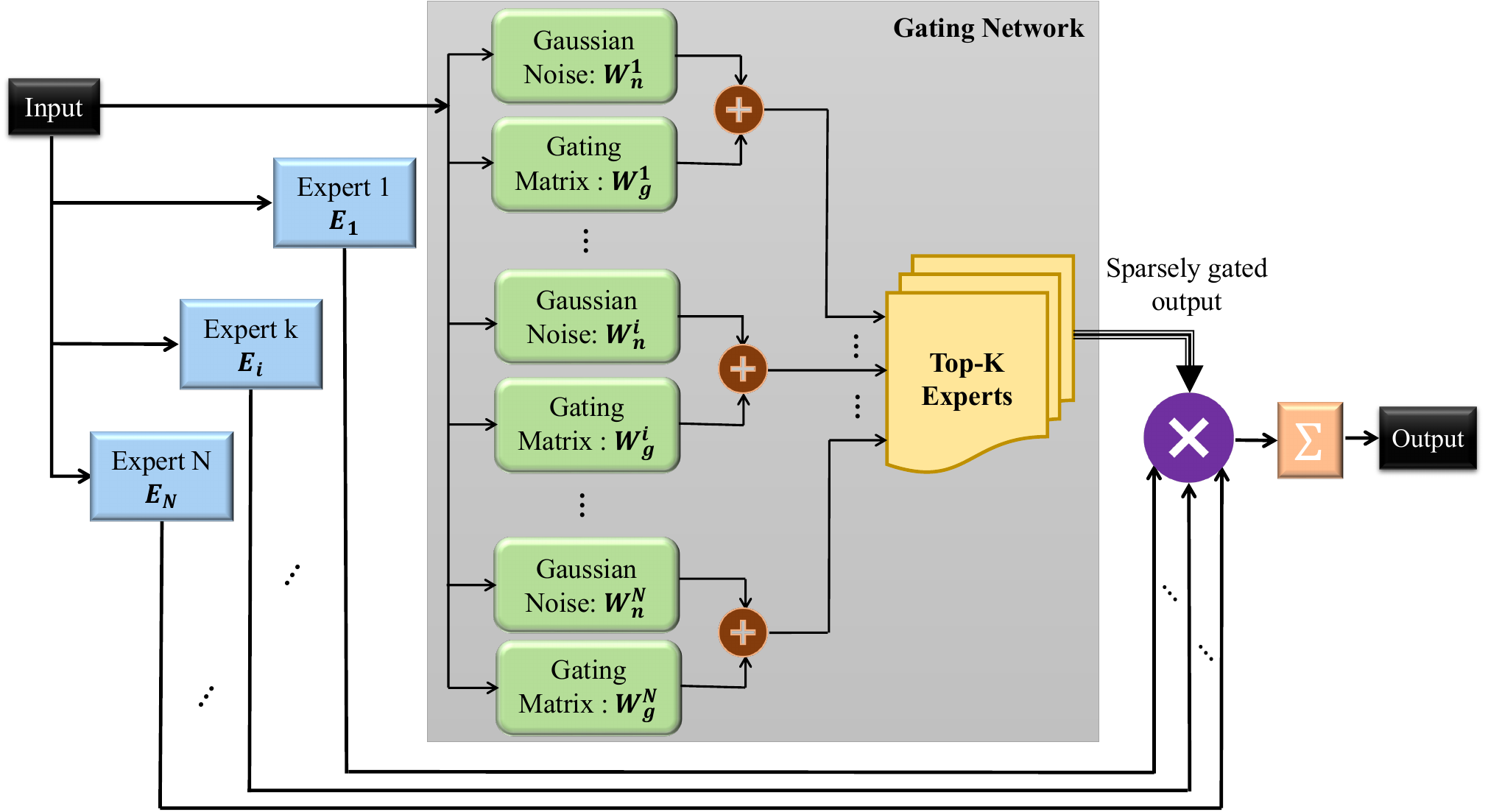}
	\caption{A Mixture of Experts (MoE) layer as described by Shazeer {\em et al.} \cite{shazeer2017}.}
	\label{fig-moe}
\end{figure*}

\subsection{Video Relationship Detection}
Video visual relationship detection (VidVRD) is made up of multiple problems that must be resolved simultaneously: object recognition, subject recognition, and action recognition. Hence, the detection often takes the form of classifying a triplet in the form of a $\!<\text{\em subject}, \text{\em predicate}, \text{\em object}\!>$ tuple for each detected subject--object pair in a video. 

When dealing with static images, this problem is comparatively straightforward as it involves detecting such a triplet only once, using language and spatial features \cite{lu2016visual,zhu2017visual,yao2018exploring,liang2018visual,kuznetsova2020open}.  In contrast, in video the problem becomes significantly more complicated as spatio-temporal features come into play, and relationships can \textit{change} over time, cascading the difficulty of the problem \cite{shang2017video,liu2019structured,shang2021video}.

In 2017, Shang \textit{et al.} \cite{shang2017video} proposed the first approach for VidVRD, decomposing a given video into segments and stitching relationship predictions in preceding and succeeding segments through a greedy association algorithm. They also introduced the first fully annotated video visual relationship dataset \cite{shang2017video}. Later, Qian \textit{et al.} \cite{qian2019video} proposed tackling the problem using a fully connected spatio-temporal graph. Tsai \textit{et al.} \cite{tsai2019video} also constructed a spatio-temporal graph, but utilized Conditional Random Fields to exploit the statistical dependency between objects. 
Liu \textit{et al.} \cite{liu2020beyond} proposed a sliding-window scheme to simultaneously predict short-term and long-term relationships using different kernel sizes on object tracklets to generate sub-tracklet proposals with different durations.

Xiao \textit{et al.} \cite{XiaoSYTC2020} proposed to spatio-temporally localize the relations (predicates) in a video sequence. Their proposed approach, the Visual Relation Grounding in Videos (vRGV) first produces region proposals from each frame of a video and then learns to ground a pre-defined relation from two trajectories. A trajectory is created by connecting the consecutive bounding boxes linked to a visual entity (subject or object) across a video segment.

Wu \textit{et al.} \cite{WuWHLL21} proposed using two graph-based networks to predict the spatial–temporal relations (actions) between subjects and objects in videos. They applied a gated graph network together with a long short-term graph network to, respectively, extract spatial relations within video frames and multi-scale temporal relations between consecutive frames.

Gao \textit{et al.} \cite{gao2021video} proposed a tracklet-based visual transformer composed of a temporal-aware decoder, which performs feature interactions between tracklets and predicate embeddings for relationship detection. Zheng \textit{et al.} \cite{Zheng_2022_CVPR} also proposed the VRDFormer, in which a first module encodes a video into a sequential frame-level feature map, and a second one processes the sequential feature map in order to generate the relation instances.

Li \textit{et al.} \cite{li2021interventional} proposed a method  to address the long-tailed bias in VidVRD datasets, which results in poor generalization. Their approach, the Interventional Video Relation Detection (IVRD) applies causality-inspired intervention on the model input to decrease the effect of the spurious correlation in the training data, and therefore to enhance the robustness of the output prediction.

Cao \textit{et al.} \cite{cao2022concept} proposed using comprehensive semantic representations that are useful for knowledge transfer across relationships to solve the VidVRD problem. Their approach, the Concept-Enhanced Relation Network (CKERN) produces conceptually richer semantic representations of the detected object pairs, and then predicts the relationship based on the integration of multi-modal features.

Gao \textit{et al.} \cite{gao2022classification} proposed a classification-then-grounding approach based on the temporal bipartite graphs of the videos, where the nodes are entities and predicates, and the edges denote different semantic roles between the nodes. Their proposed approach, the Bipartite Graph model (BIG) first classifies all of the nodes and edges of the graph (classification), and then localizes the temporal location of each relation instance (grounding).

Chen \textit{et al.} \cite{chen2021social} introduced a compositional encoding for VidVRD. Their proposed approach, the Social Fabric Encoding (SFE) encodes a pair of object ``tubelets'' as a composition of interaction primitives. Learning these primitives, the resulting representation is used to localize and classify relationships from co-occurring objects.

More recently, Shang \textit{et al.} \cite{shang2021video} modified their earlier approach \cite{shang2017video}, whereby they proposed an iterative relation inference that exploits the inter-dependency of relation components (subject/objects and predicates) for better visual relation recognition. To achieve this, they created three preferential predictors with learnable tensors alongside the normal visual predictors to model the inter-dependency relationship between subjects/objects and predicate classes \cite{shang2021video}. Hence, each relational component has three classifiers, and each of them consists of a visual predictor and a preferential predictor. The visual predictor is a deep neural network for recognizing the visual patterns of subject/object and predicate, whereas the preferential predictor refines the prediction of one variable (subject, object, or predicate) conditioned on the values of the other two \cite{shang2021video}. Following a similar architecture to their initial work \cite{shang2021video}, the authors take a sliding time window and generate object tracklet proposals as the detected entities, and then predict associated relation triplets.

Due to its lightweight network architecture, modularity, and state-of-the-art performance, we have chosen the work by Shang \textit{et al.} \cite{shang2021video} as the basis for the expert in the architecture proposed in this paper.

\subsection{Mixture of Experts}
\label{sect:moe}

Proposed more than three decades ago by Jacobs \textit{et al.} \cite{jacobs1991adaptive}, the mixture of experts (MoE) architecture has been applied to problems including the modeling of task relationships \cite{ma2018modeling}, increasing network breadth and depth \cite{wang2020deep}, multi-modal generative models \cite{shi2019variational}, and volunteer computing \cite{ryabinin2020towards}.

In 2017, Shazeer \textit{et al.} proposed a new general purpose neural network component: the Sparsely-Gated Mixture-of-Experts (MoE) Layer \cite{shazeer2017}, consisting of a number of experts, each a simple feed-forward neural network, together with a trainable gating network which selects a sparse subset of the experts to be trained on each given input \cite{shazeer2017}. The gating network essentially determines which of the experts are best suited to a given type of input.  All parts of the
network, both gating and experts, are trained jointly by back-propagation \cite{shazeer2017}. In their paper, Shazeer \textit{et al.} applied their technique, illustrated in Figure~\ref{fig-moe}, to language modeling and machine translation.

Riquelme \textit{et al.} also proposed a Vision Mixture of Expert (V-MoE) \cite{RiquelmePMNJPKH21} for image classification. V-MoE replaces a subset of feedforward layers in a vision transformer with sparse MoE layers, where each image patch is ``routed'' to a subset of ``experts''. This halves the computation consumption at inference while performing equally well as the state-of-the-art.  

It has been observed \cite{shazeer2017, eigen2013learning, wang2020deep} that a gating network is inclined to converge to a state where it produces
large weights for the same few experts regardless of input, a phenomenon very much analogous to the problems encountered with self-organized maps \cite{kohonen1990self} (essentially a very flat single-layer network) from pattern recognition. This imbalance becomes self-reinforcing / self-perpetuating, as the favored experts
are trained more frequently / more rapidly and thus are even more likely to be selected by the gating network \cite{shazeer2017}. To address this problem, Shazeer \textit{et al.}\ \cite{shazeer2017} defined the importance of an expert relative to a batch of
training samples to be the batch-wise sum of the gate values for that expert.

In the following section we will present our proposed architecture, which applies a sparsely gated mixture of experts \cite{shazeer2017} to video visual relationship detection \cite{shang2021video}.

\section{Architecture}
\label{sect:arch}
Figure~\ref{fig:moe-VRD} illustrates our proposed architecture and its main components.

\subsection{Sparsely-Gated Mixture of Experts}
\label{sgmoe}

The MoE consists of a set of $N$ expert networks $\text{E}_1,..., \text{E}_{N}$ and one gating network, $\text{G}$, whose output is a sparse binary $N$-dimensional vector. The experts are themselves identical feed-forward neural networks, each with their own parameters.  Since our interest in this paper is the MoE concept, for the individual experts we have adopt the baseline state-of-the-art approach of \cite{shang2021video} shown in Figure~\ref{fig-moe}. 

Given an input $x$, the output of the $i$th expert's function is denoted as ${\text{E}}_i(x)$.  These $N$ outputs are combined in the MoE layer as
\begin{equation}
	y = \sum_{i=1}^N {\text{G}}(x)_i \cdot {\text{E}}_i(x),
\end{equation}
where ${\text{G}}(x)_i$ represents the output of the gating network.  
The sparsity in computation, one of the key strengths of the MoE approach, is realized by the explicit sparsity of the gating output.  That is,
\begin{equation}
	{\text{G}}(x)_i = 0 \qquad \text{for most $i$,}
\end{equation}
such that whenever ${\text{G}}(x)_i = 0$ the corresponding expert ${{E}}_i$ is not invoked (the particular input $x$ is not fed-forward into the network representing expert ${{E}}_i$).

There is significant flexibility in the choice of gating function.  In this paper we adopt the absolute simplest case --- a single--layer gating function, with more capable multi-layer generalizations to be considered as future work.  The gating output ${\text{G}}(x)$ is computed as
\begin{equation}
	{\text{G}}(x) = \text{Softmax} \big(\text{top}_{K} \big(W^i_g \cdot x+ \text{N}_{g}(W^i_n \cdot x)\big)\big),
\end{equation}
where $\text{top}_{K}$ selects the $K$ largest values (the best experts),
and $W^i_g$ and $W^i_n$ are trainable gating and noise weight matrices, respectively, which are parametrized for each expert $i$.

The number of samples sent to the gating layer is discrete, and therefore not applicable to back-propagation, however the inclusion of the noise term $\text{N}_{g}(x)$ allows for a smooth estimate of the number of samples used for each expert in each batch, thus allowing for the back-propagation of gradients.  The noise function is defined as
\begin{equation}
	\text{N}_{g}(x) = \text{StandardNormal}() \cdot \text{Softplus}(x),
\end{equation}
where $\text{Softplus}(x)=\frac{1}{\beta}\log(1+ \beta x)$ is a smooth approximation of the $\text{ReLU}$ function to constrain the output to be positive.

An importance term is considered in the overall loss to address imbalances resulting from the \textit{self-reinforcing effect} \cite{lu2021self}, which occurs when certain favoured experts are trained more rapidly and thus are selected even more by the gating network.  The importance loss is
\begin{equation}
	\label{limp}
	\text{L}_\text{importance}(x) = \alpha \big( \text{CV}(g) + \text{CV}(l)\big)\big),
\end{equation}
where $\alpha$ is a hand-tuned scaling factor, $g$ is the batch-wise sum of gate values
\begin{equation}
	g = \sum_{x \in B} \text{G}(x)
\end{equation}
over batch $B$, and 
\begin{equation}
	l = \sum_{x \in B, \text{G}(x)>0} \text{G}(x)
\end{equation}
represents the load, summed over the positive gate values. 

We applied $\text{CV}(\cdot)$, the coefficient of variation,
\begin{equation}
	\text{CV}(x) = \frac{\text{var}(x)}{\bigl(\text{mean}(x)\bigr)^2 + \epsilon}
\end{equation}
as additional loss terms in (\ref{limp}) which encourage experts to have a more balanced (equal) importance \cite{shazeer2017}.

\subsection{Visual relationship detection}
\begin{figure*}[tp]
	\centering
	\includegraphics[width=\textwidth]{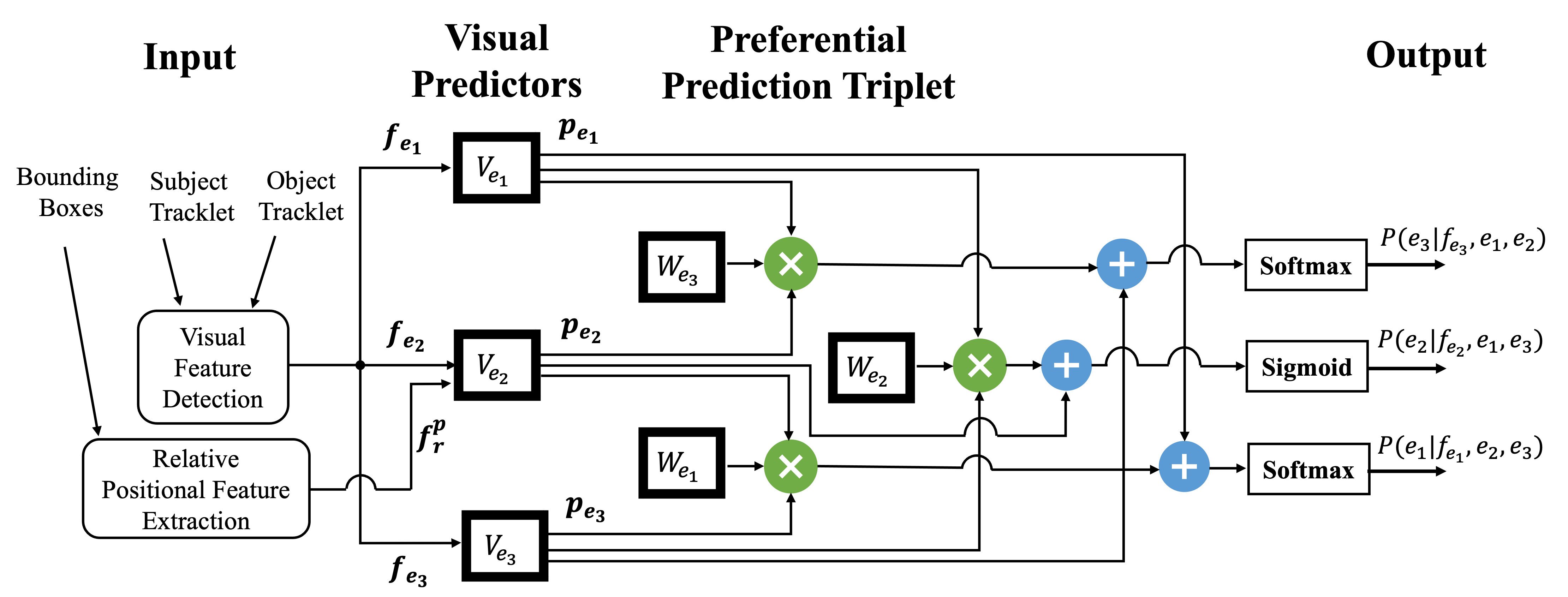}
	\caption{Visual relationship detection framework proposed by \cite{shang2021video}, which is used as the basis of our expert.}
	\label{fig:VRD}
\end{figure*}

To address the problem of visual relationship detection in video sequences, we take inspiration from VidVRD-II \cite{shang2021video}, which learns relationship detection using iterative inference shown in Figure~\ref{fig:VRD}. 

VidVRD-II assumes a set of three entities $\mathbb{E}=[e_1, e_2, e_3]$ representing subject $e_1$, predicate $e_2$ and object $e_3$, and their corresponding features $f_{e_{1}}$, $f_{e_{2}}$, $f_{e_{3}}$, which builds the language triplet $<\text{\em subject}, \text{\em predicate}, \text{\em object}>$, and models the problem of video visual relationship detection as the joint probability
\begin{equation}\label{eq:joint-prob}
	{P}(<e_1, e_2, e_3>|f_{e_{1}}, f_{e_{2}}, f_{e_{3}}).
\end{equation}
This joint probability can be factorized as three conditional probabilities
\begin{equation}\label{eq:cascade}
	{P}(e_{1}|f_{e_{1}}, e_{2}, e_{3}) {P}(e_{2}|f_{e_{2}}, e_{1}, e_{3}) {P}(e_{3}|f_{e_{3}}, e_{1}, e_{2}) ,
\end{equation}
which aids in inference when there is ambiguous visual information, since the classes of any two components imply a preference over the class of the third.

Each conditional probability of (\ref{eq:cascade}) is modelled by a classifier consisting of a visual predictor and a preferential predictor also shown in Figure~\ref{fig:VRD}. The visual predictor simply learns visual patterns of the subject $e_1$, predicate $e_2$, and object $e_3$, using a deep neural network. The preferential predictor applies learnable dependency tensors to refine the prediction of one variable conditioned on the values of the other two:
\begin{equation}\label{eq:prefpred}
	e_{pr}=\left\{
	\begin{array}{@{}ll@{}}
		{P}(e_{1}|f_{e_{1}}, e_{2}, e_{3})= \Phi(V_{e_1} \cdot f_{e_{1}}) + p_{e_2} \cdot {W}_{e_1} \cdot p_{e_3} \\
		
		{P}(e_{2}|f_{e_{2}}, e_{1}, e_{3})= \Phi({V}_{e_2} \cdot f_{e_{2}}) + p_{e_1} \cdot {W}_{e_2} \cdot p_{e_3} \\
		
		{P}(e_{3}|f_{e_{3}}, e_{1}, e_{2})= \Phi({V}_{e_3} \cdot f_{e_{3}}) + p_{e_1} \cdot {W}_{e_3} \cdot p_{e_2}
	\end{array}\right.
\end{equation} 
Here $e_{pr}$ is the conditional probability vector of three entities, and ${V} = [{V}_{e_1}, {V}_{e_2}, {V}_{e_3}]$ are the learnable weights of the visual predictors. In our case study, the weights of the subject and object classifiers are shared, thus ${V}_{e_1} = {V}_{e_3}$.  The paper of \cite{shang2021video} applied the nonlinearity $\Phi$ to the entire expression of (\ref{eq:prefpred}), as in 
\begin{equation}
	{P}(e_{i}|f_{e_{i}}, e_{j}, e_{k})= \Phi({V}_{e_i} \cdot f_{e_{i}} + p_{e_j} \cdot {W}_{e_i} \cdot p_{e_k})
\end{equation} 
However actual implementations applied $\Phi$ to the first term only, as in (\ref{eq:prefpred}), a convention which we have preserved for consistency.
The weights $[{W}_{e_1}, {W}_{e_2}, {W}_{e_3}]$ model the dependency of one class over the other two, separately parametrized for each classifier. $\Phi$ represents the nonlinear activation, here implemented by a $\text{Softmax}$ function for the subject and object classes, and a $\text{Sigmoid}$ function for the predicate class.

To design our architecture, we apply the MoE concept of Section~\ref{sgmoe} by incorporating VidVRD-II into the MoE framework, giving rise to the combined MoE-VRD shown in Figure~\ref{fig:moe-VRD}.  The object tracklet proposals are extracted and fed into the relation prediction experts.  Based on the gating layer, the top $K$ experts are chosen. Upon the completion of the forward pass, back-propagation is applied so the gradients back-propagate through the gating network and the selected experts.

\subsubsection{Object Tracklet Proposals}
We use Seq-NMS \cite{han2016seq} to generate object tracklet proposals as a pre-processing step to use as inputs to the relational classifier experts. For frame-level object detection, a Faster-RCNN with an Inception-ResNet foundation \cite{szegedy2017inception} is pretrained on the Open Images dataset \cite{kuznetsova2020open}.

The model serves as a suitably generic object detector \cite{szegedy2017inception,shang2021video}. The bounding boxes and corresponding region features are extracted, after which Seq-NMS generates a compact set of object tracklets,
which form the inputs to the expert neural networks.

\subsubsection{Feature Extraction}
\begin{figure*}[!htb]
	\centering
	\includegraphics[scale=.47]{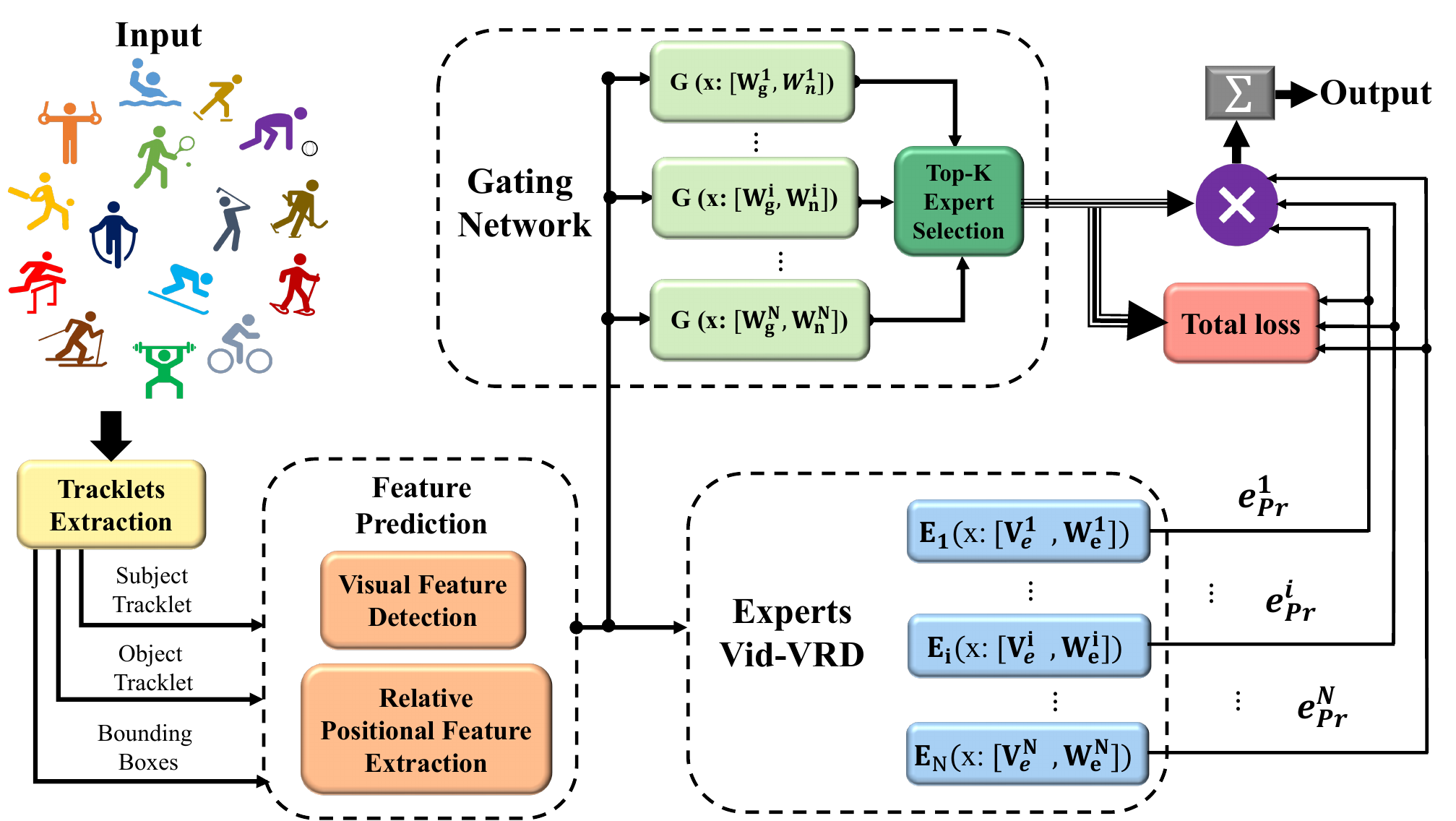}
	\caption{An illustration of the MoE-VRD architecture proposed in this article. Raw RGB images are taken as input; for each given image frame the subject and object tracklets are extracted and given to the feature extraction network together with bounding box information in order to generate visual and relative positional features representing all three entities: subject, predicate and object. The visual and positional features are applied as the input to our experts and gating networks. Every expert outputs a score corresponding to each entity, which represents both visual and preferential predictions. The gating network outputs a sparsely gated vector, which evaluates each expert's learning. Selecting the top $K$ experts, the sum-product of the sparsely gated expert scores is calculated and represented as the output of our MoE-VRD architecture.}
	\label{fig:moe-VRD}
\end{figure*}

Applying the object tracklet proposals, we generate two types of features:  \textit{Visual Features} and \textit{Relative Positional Features}, shown in Figure~\ref{fig:moe-VRD}. 

To generate the visual features $f$ of (\ref{eq:joint-prob}) -- (\ref{eq:cascade}), the bounding boxes are applied to extract the pretrained deep visual features of the subject and object entities, and the predicate's visual feature is computed through a concatenation of the subject and object visual feature vectors.

In addition to the visual features, we extract a relative positional feature to represent the spatio-temporal relationship between the entities.  For each pair of object tracklets, the algorithm computes the relative distance between the subject and object by encoding the spatial and temporal relative positional feature $f_{r}^{p}$:
\begin{equation}
	\begin{split}
		f_{r}^{p} = \Bigg[ 
		\frac{x^p_{e_{1}} - x^p_{e_{3}}}{x^p_{e_{3}}}, 
		\frac{y^p_{e_{1}} - y^p_{e_{3}}}{y^p_{e_{3}}}, 
		\log \frac{w^p_{e_{1}}}{w^p_{e_{3}}}, 
		\log \frac{h^p_{e_{1}}}{h^p_{e_{3}}}, 
		\log \frac{w^p_{e_{1}} h^p_{e_{1}}}{w^p_{e_{3}} h^p_{e_{3}}}, 
		\frac{t^p_{e_{1}} - t^p_{e_{3}}}{30} 
		\Bigg],
	\end{split}
\end{equation}
where $p \in [b, e]$ represents the beginning or ending bounding box, characterized by  coordinates $(x, y)$, width $w$, height $h$, and time $t$ for subject $e_{1}$ and object $e_{3}$. A feed-forward network is used to fuse the subject's and object's visual features $f_{e_{1}}$, $f_{e_{3}}$ with the relative positional features of the beginning and ending bounding boxes $f_{r}^{b}$,  $f_{r}^{e}$, where the relative positional feature $f_{r}^{p}$ provides the expert with additional information to recognize visual relationships. 

In summary, each encapsulated expert consists of an object predictor, a subject predictor, and a predicate predictor --- each of which is a basic feed-forward network, allowing for a set of modestly-sized, nimble experts to speed up training and inference, when compared to an equivalent single monolithic network.


%
%


\section{Experiments and Results}
\label{experiments}

In order to properly assess the improvements offered by our proposed framework and to make a fair comparison with the state-of-the-art, we conduct our experiments using a similar experimental setup and the same datasets as used by the iterative inference approach proposed by Shang \textit{et al.} \cite{shang2021video}.

\subsection{Datasets} Our experiments are conducted using two VidVRD benchmark datasets: ImageNet-VidVRD \cite{shang2017video} and VidOR \cite{shang2019annotating,thomee2016yfcc100m}. 
The ImageNet-VidVRD is the first dataset for video visual relation detection, created by Shang \textit{et al.} \cite{shang2017video}. It consists of 1,000 videos collected from ILSVRC2016-VID \cite{russakovsky2015imagenet}. These videos are manually annotated with video relation instances \cite{shang2017video,shang2021video}. VidOR is a recently-released large-scale benchmark also compiled by Shang \textit{et al.} \cite{shang2019annotating}, which contains 10,000 social media videos from YFCC-100M \cite{thomee2016yfcc100m}.

\subsection{Evaluation metrics} 
In object detection there are two related problems to be solved: localization and classification. Localization determines the location of an object (e.g., its bounding box), whereas classification infers the object's identity.

For object detection tasks, it is standard to calculate precision and recall based on a given threshold on the IoU (Intersection over Union), which measures the fractional overlap between predicted and true bounding boxes.  If the IoU result for a predicted bounding box exceeds the threshold, then the prediction is classified as a true positive, otherwise it is a false positive.

We can thus define metrics, such as Recall@50, implying an evaluation of the recall metric based on an IoU threshold of 50\% (i.e., allowing bounding boxes to be only 50\% overlapping).  Other metrics, such as Recall@100, Precision@10 etc.\ are equivalently defined.

We designed and implemented our evaluation metrics consistent with those in previous works \cite{shang2017video, shang2019annotating, shang2021video}. To this end, we calculate how many ground truth relation instances are detected by the mixture of experts in each testing video. Metrics are divided into two categories: relation tagging and relation detection.

Relation tagging focuses on the precision of the relation triplet without considering the precision of its spatio-temporal location in the video. In other words, relation tagging simply checks whether the relationship was detected properly, but not whether it was detected with any accuracy in time or space. 
The tagging performance is evaluated by Precision@1 (P@1), Precision@5 (P@5), and Precision@10 (P@10).

In contrast, relation detection measures the precision of both the relation triplet and the corresponding subject/object trajectories for every detected relation instance \cite{shang2017video}. Relation detection of instances are considered to be correct only if they match the ground truth relation instance, and the voluminal Intersection-over-Union (vIoU) of the trajectories of subject and object are both larger than a pre-defined threshold. 
The vIoU threshold is set to 50\% in order to address the objectives of our experiments. The detection performance is measured using the Mean Average Precision (mAP), Recall@50 (R@50), and Recall@100 (R@100). 

All experiments are run ten times with different random seeds for each expert, and the mean plus/minus standard-deviation scores are reported.

\subsection{Single Expert Performance}
To ensure the proper functioning of the proposed mixture of experts, clearly we must first validate the performance of a single expert.  The performance of the MoE framework consisting of only a single expert ($N=1$, and thus necessarily ${K} = 1$) during training and testing should be essentially unchanged from the published performance of the underlying expert \cite{shang2017video,shang2021video}. 

In the multi-expert architecture, at every iteration the gating layer outputs a sparse vector selecting $K$ experts.  If there is only a single expert, then the gating layer is essentially irrelevant, simply selecting the same one expert every time, and the resulting performance should be the same as if we had simply run the model built by Shang \textit{et al.} without any adjustments \cite{shang2021video}.

\begin{table*}
    \small
    \centering
    \begin{tabularx}{\textwidth}{Xlllllll}
        \toprule
        \multicolumn{2}{c}{ImageNet-VidVRD Dataset} & \multicolumn{3}{c}{\textbf{Relation detection}} & \multicolumn{3}{c}{\textbf{Relation tagging}} \\
        \cmidrule(lr){3-5} \cmidrule(lr){6-8}
        & & \textbf{mAP} & \textbf{R@50} & \textbf{R@100} & \textbf{P@1} & \textbf{P@5} & \textbf{P@10} \\
        \midrule
        VidVRD-II\cite{shang2021video} & & $29.37 \pm 0.40$ & $19.63 \pm 0.19$ & $22.92 \pm 0.48$ & $70.40 \pm 1.53$ & $53.88 \pm 0.31$ & $40.16 \pm 0.70$ \\
        \addlinespace
        \textbf{MoE-VRD} ($N = 1$) & & $29.49 \pm 0.32$ & $18.65 \pm 0.2$ & $22.71 \pm 0.33$ & $68.50 \pm 1.81$ & $54.70 \pm 0.23$ & $39.50 \pm 0.86$ \\
        \bottomrule
    \end{tabularx}
    
    \vspace*{0.1in}
    
    \begin{tabularx}{\textwidth}{Xlllllll}
        \toprule
        \multicolumn{2}{c}{VidOR Dataset} & \multicolumn{3}{c}{\textbf{Relation detection}} & \multicolumn{3}{c}{\textbf{Relation tagging}} \\
        \cmidrule(lr){3-5} \cmidrule(lr){6-8}
        & & \textbf{mAP} & \textbf{R@50} & \textbf{R@100} & \textbf{P@1} & \textbf{P@5} & \textbf{P@10} \\
        \midrule
        VidVRD-II\cite{shang2021video} & & $8.65 \pm 0.11$ & $8.59 \pm 0.11$ & $10.69 \pm 0.08$ & $57.40 \pm 0.57$ & $44.54 \pm 0.68$ & $33.30 \pm 0.31$ \\
        \addlinespace
        \textbf{MoE-VRD} ($N = 1$) & & $7.98 \pm 0.16$ & $8.23 \pm 0.28$ & $11.02 \pm 0.14$ & $57.53 \pm 0.43$ & $43.12 \pm 0.54$ & $32.89 \pm 0.24$ \\
        \bottomrule
    \end{tabularx}
    
    \caption{Comparisons of a single expert with the method of by Shang \textit{et al.} \cite{shang2021video} on the ImageNet-VidVRD Dataset \cite{shang2017video} (top) and on the VidOR Dataset \cite{shang2019annotating} (bottom).  In both cases, the expert performs essentially equivalently to that of \cite{shang2021video}.}
    \label{tab:k1_results}
\end{table*}

We have shown the results of the state-of-the-art approaches in Table~\ref{tab:k1_results}, illustrating a comparison between the relation detection and relation tagging results of Shang \textit{et al.}'s VidVRD-II \cite{shang2021video} and our single-expert ($N = 1$) MoE-VRD architecture over the ImageNet-VRD and VidOR datasets.  Both approaches perform quite similarly, validating that the MoE-VRF framework is not interfering with the operation of the underlying expert, allowing us to generalize to multiple experts, next.

\begin{table*}[htb]
    \footnotesize
    \setlength{\tabcolsep}{3pt} 
    \centering
    \begin{tabularx}{\textwidth}{Xlllllll}
        \toprule
        \multicolumn{2}{c}{ImageNet-VidVRD Dataset} & \multicolumn{3}{c}{\textbf{Relation detection}} & \multicolumn{3}{c}{\textbf{Relation tagging}} \\
        \cmidrule(lr){3-5} \cmidrule(lr){6-8}
        & & \textbf{mAP} & \textbf{R@50} & \textbf{R@100} & \textbf{P@1} & \textbf{P@5} & \textbf{P@10} \\
        \midrule
        VidVRD\cite{shang2017video} & & $8.58$ & $5.54$ & $6.37$ & $43.00$ & $28.90$ & $20.80$ \\
        GSTEG\cite{tsai2019video} & & $9.52$ & $7.05$ & $7.67$ & $51.50$ & $39.50$ & $28.23$ \\
        VRD-GCN\cite{qian2019video} & & $14.23$ & $7.43$ & $8.75$ & $59.50$ & $40.50$ & $27.85$ \\
        3DRN\cite{cao20213} & & $14.68$ & $5.53$ & $6.39$ & $57.89$ & $41.80$ & $29.15$ \\
        VRD-STGC\cite{liu2019structured} & & $18.38$ & $11.21$ & $13.69$ & $60.00$ & $43.10$ & $32.24$ \\
        SFE\cite{chen2021social} & & $20.08$ & $13.73$ & $16.88$ & $62.50$ & $49.20$ & $38.45$ \\
        IVRD\cite{li2021interventional} & & $22.97$ & $12.40$ & $14.46$ & $68.83$ & $49.87$ & $35.57$ \\
        BIG-C\cite{gao2022classification} & & $26.08$ & $14.10$ & $16.25$ & $73.00$ & $55.10$ & $40.00$ \\
        CKERN\cite{cao2022concept} & & $-$ & $-$ & $-$ & $74.50$ & $55.59$ & $41.34$ \\
        VidVRD-II\cite{shang2021video} & & $29.37 \pm 0.40$ & $19.63 \pm 0.19$ & $22.92 \pm 0.48$ & $70.40 \pm 1.53$ & $53.88 \pm 0.31$ & $40.16 \pm 0.70$ \\
        \textbf{MoE-VRD} ($\text{K} = 2$) & & $\mathbf{33.02 \pm 0.23}$ & $\mathbf{22.77 \pm 0.28}$ & $\mathbf{24.20 \pm 0.22}$ & $\mathbf{74.12 \pm 1.44}$ & $\mathbf{56.47 \pm 0.17}$ & $\mathbf{42.05 \pm 0.92}$ \\
        \bottomrule
    \end{tabularx}
    \caption{Performance of our proposed MoE-VRD with $K = 2$ and a total of $N=10$ experts, in comparison with state-of-the-art approaches on the ImageNet-VidVRD dataset \cite{shang2017video}. For \emph{every} criterion the proposed MoE-VRD outperforms all other approaches. The substantial increase in performance stems unambiguously from the mixture-of-experts approach, since our expert on its own is no better than the method in VidVRD-II, as was shown in Table~\ref{tab:k1_results}. "$-$" indicates that no corresponding results were reported.}
    \label{tab:overall_results_imagenet}
\end{table*}

\begin{table*}[ht!]
    \footnotesize
    \setlength{\tabcolsep}{3pt} 
    \centering
    \begin{tabularx}{\textwidth}{Xlllllll}
        \toprule
        \multicolumn{2}{c}{VidOR Dataset} & \multicolumn{3}{c}{\textbf{Relation detection}} & \multicolumn{3}{c}{\textbf{Relation tagging}} \\
        \cmidrule(lr){3-5} \cmidrule(lr){6-8}
        & & \textbf{mAP} & \textbf{R@50} & \textbf{R@100} & \textbf{P@1} & \textbf{P@5} & \textbf{P@10} \\
        \midrule
        3DRN\cite{cao20213} & & $2.47$ & $1.58$ & $1.85$ & $33.05$ & $35.27$ & $-$ \\
        VRD-STGC\cite{liu2019structured} & & $6.85$ & $8.21$ & $9.90$ & $48.92$ & $36.78$ & $-$ \\
        IVRD\cite{li2021interventional} & & $7.42$ & $7.36$ & $9.41$ & $53.40$ & $42.70$ & $-$ \\
        CKERN\cite{cao2022concept} & & $-$ & $-$ & $-$ & $58.80$ & $46.07$ & $34.29$ \\
        BIG\cite{gao2022classification} & & $8.54$ & $8.03$ & $10.04$ & $64.42$ & $51.80$ & $40.96$ \\
        Ens-5\cite{gao2021video} & & $9.48$ & $8.56$ & $10.43$ & $63.46$ & $54.07$ & $\mathbf{41.94}$ \\
        SFE\cite{chen2021social} & & $\mathbf{11.21}$ & $\mathbf{9.99}$ & $\mathbf{11.94}$ & $\mathbf{68.86}$ & $\mathbf{55.16}$ & $-$ \\
        VidVRD-II\cite{shang2021video} & & $8.65 \pm 0.11$ & $8.59 \pm 0.11$ & $10.69 \pm 0.08$ & $57.40 \pm 0.57$ & $44.54 \pm 0.68$ & $33.30 \pm 0.31$ \\
        \textbf{MoE-VRD}($\text{K} = 2$) & & $9.44 \pm 0.21$ & $9.54 \pm 0.13$ & $11.51 \pm 0.31$ & $58.92 \pm 0.67$ & $45.11 \pm 0.19$ & $34.85 \pm 0.22$ \\
        \bottomrule
    \end{tabularx}
    \captionsetup{font=footnotesize} 
    \caption{A comparison with the state-of-the-art, as in Table~\ref{tab:overall_results_imagenet}, but here on the VidOR dataset \cite{shang2019annotating}.}
    \label{tab:overall_results_vidor}
\end{table*}

\begin{figure*}[!htb]
	\centering
	\includegraphics[scale=.5]{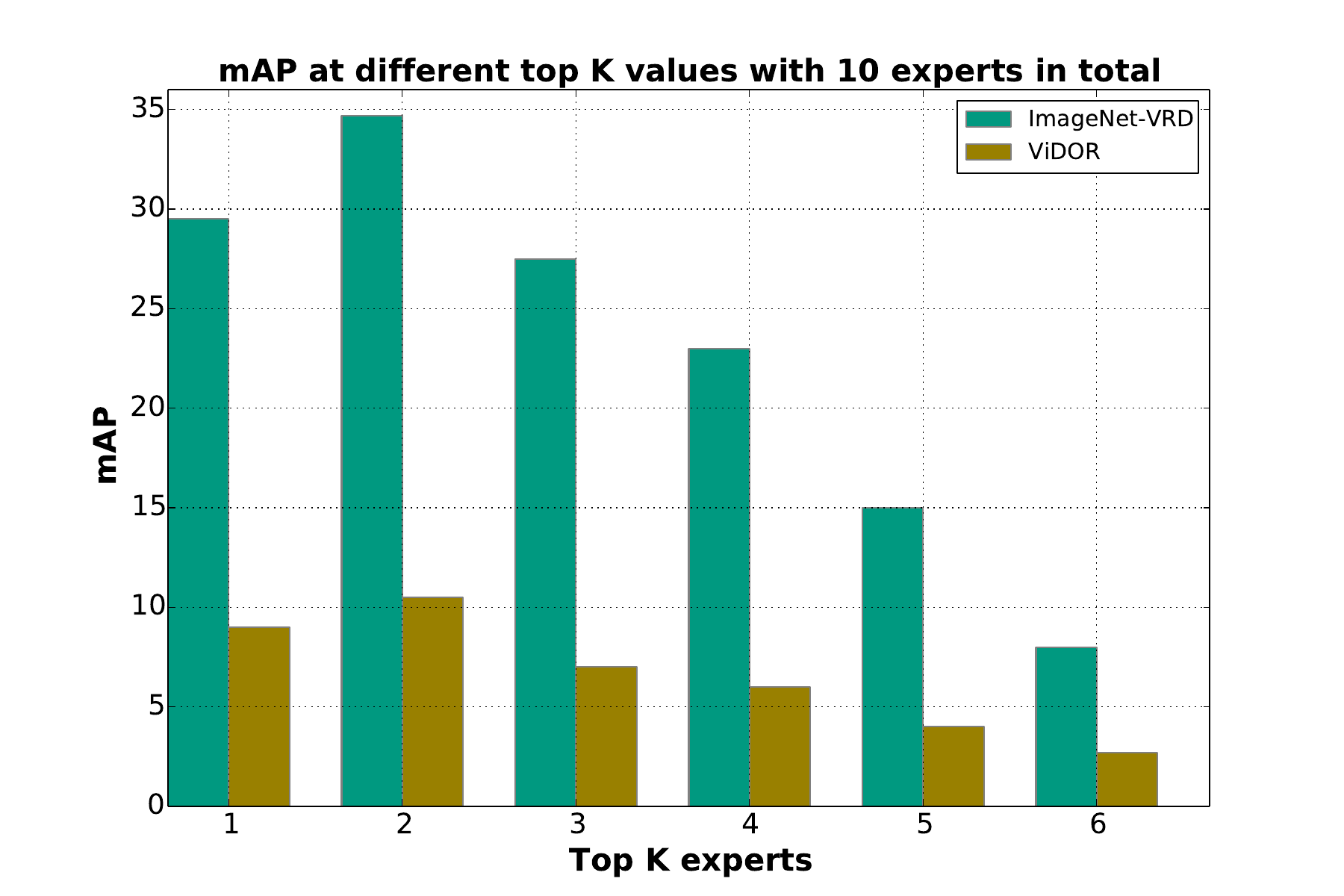}
	\caption{mAP of the MoE-VRD approach having $N=10$ experts, as a function of $K$ during training. Note that performance drops after $K=2$; due to the averaging nature of the architecture before the final output, such that well-performing experts may become drowned out by more poorly performing peers if $K$ is set too large.}
	\label{fig:k_results}
\end{figure*}

\subsection{Multi-expert Performance}

We now wish to evaluate the performance of our proposed MoE-VRD architecture when more than one expert ($N>1$) is at play. We evaluate our approach on the VidOR dataset \cite{shang2019annotating} and ImageNet-VRD dataset \cite{shang2017video}.

We evaluate our work against recent representative approaches:
\begin{itemize}
	\item \textbf{VidVRD-II} \cite{shang2021video}, which builds upon the same authors' work in \cite{shang2017video}. It uses an iterative inference approach to video relationship detection.
	\item \textbf{GSTEG} \cite{tsai2019video}, which constructs a fully-connected spatio-temporal graph for relation inference.
	\item \textbf{VRD-GCN} \cite{qian2019video}, which builds a model that can take advantage of spatial-temporal contextual cues to make better predictions on objects as well as their dynamic relationships.
	\item \textbf{VRD-STGC} \cite{liu2020beyond}, which proposes a novel sliding-window scheme to simultaneously predict short-term and long-term relationships \cite{liu2020beyond}, and extracts spatio-temporal features.
	\item \textbf{3DRN} \cite{cao20213}, which develops a 3-D CNN to learn the visual features for relation recognition in an end-to-end manner.
	\item \textbf{IVRD} \cite{li2021interventional}, which proposes a causality-inspired intervention on the model input to improve prediction robustness.
	\item \textbf{CKERN} \cite{cao2022concept}, which generates comprehensive semantic representations by incorporating retrieved concepts with local semantics.
	\item \textbf{BIG} \cite{gao2022classification}, which proposes a classification-then-grounding approach based on temporal bipartite graphs.
	\item \textbf{Ens-5} \cite{gao2021video}, which proposes a tracklet-based visual Transformer composed of a temporal-aware decoder.
	\item \textbf{SFE} \cite{chen2021social}, which proposes encoding the representation of a pair of objects as a composition of interaction primitives. 
\end{itemize}
For the ImageNet-VRD dataset \cite{shang2017video} we compare to all ten of these methods; for the VidOR dataset \cite{shang2019annotating} we compare to eight of the preceding methods, due to the choice of results reported in the respective papers.

Table~\ref{tab:overall_results_imagenet} shows the results, comparing our proposed MoE architecture with all ten other methods on the ImageNet-VRD dataset \cite{shang2017video}.  The proposed MoE-VRD performs significantly better, in every metric, than any method tested, including the most recent state of the art. The large margin of improvement stems from having a gating function that allows experts to be trained quite separately on different sorts of inputs, leading to a degree of robustness due to heterogeneity, which is very difficult for single large monolithic networks.

Similar to Table~\ref{tab:overall_results_imagenet}, Table~\ref{tab:overall_results_vidor} now shows the comparative results on the VidOR dataset \cite{shang2019annotating}.  Our proposed MoE-VRD still exhibits superior performance in every metric when compared to most of the state of the art approaches, although by a lesser margin than in Table~\ref{tab:overall_results_imagenet}, likely due to the increased diversity of the VidOR dataset  \cite{shang2019annotating}, and the related naivety or limitation of the Moe-VRD in using a set of identical experts.  The creation of heterogeneous or differently-specialized experts is a subject for future research.

BIG \cite{gao2022classification}, Ens-5 \cite{gao2021video}, and SFE\cite{chen2021social} do outperform the proposed MoE-VRD for one or more metrics in Table~\ref{tab:overall_results_vidor}, although for the Relation Detection assessment the MoE-VRD is highly competitive, outperforming BIG and Ens-5.  In any event, a universal improvement on every possible dataset and/or metric is not to be expected, and the impressive results of SFE in Table~\ref{tab:overall_results_vidor} are, for example, significantly less impressive in its rather lackluster performance, relative to MoE-VRD, in Table~\ref{tab:overall_results_imagenet}.

\subsection{Ablation study} 
Really the only aspect of the proposed architecture which can be removed, via ablation, is the collection of experts.  That is, our ablation study assesses performance as a function of the number of  ``top experts'' $K$ chosen by the gating function for each input.  The total number of experts was fixed to $N=10$, since $N>K$ needs to be large enough to test a meaningful range of $K$, at the same time increasing $N$ far past 10 leads to challenges in network memory requirements and training reliability.  The results of the multi-expert experiments are presented in Figure~\ref{fig:k_results}, plotting mAP $K$, ablating $K$ from 6 down to 1.

The best MoE results are achieved when we select the two top experts ($K=2$) for each input, such that the performance drops with increased $K$ for both datasets.  Note that a low optimum value of $K$ is an asset, not a liability, in that a small $K$ implies a modest computational complexity, since only $K$ experts are actually engaged for any given input.

\section{Discussion \& Conclusions}
\label{discuss}

The problem of video-based visual relationship detection (Vid-VRD) is relatively new compared to static image-based visual relationship detection. The spatio-temporal dimensions in the video domain cascade the difficulty of the problem, given the far greater data volumes and the ability for relationships to change over time. There have been a few approaches to address this problem, however they uniformly rely on monolithic neural networks \cite{tsai2019video,qian2019video,liu2019structured}.

In this work, we developed a new framework, the MoE-VRD, based on a mixture of experts approach.  MoE-VRD is developed by encapsulating a Vid-VRD framework \cite{shang2021video} into an expert within a sparsely gated mixture of experts architecture.

Our proposed approach to video visual relationship detection also addresses limitations in computing power and distributed computation, which arises from the limited capacity of neural networks to absorb information due to the limitations in network size (number of parameters) in comparatively blunt architectures based on a single, monolithic network.

We have observed that the performance of the network drops when we select more than two experts ($K>2$). We believe that this stems from the averaging operation, which acts prior to the final output, resulting in well-performing experts being increasingly drowned out by those experts having inferior performance. Studying this effect more carefully, and in other settings, is one subject for future work.

We achieved highly promising results from MoE-VRD based on experiments on two different datasets, ImageNet-VRD \cite{shang2017video} and VidOR \cite{shang2019annotating}.  Our MoE-VRD outperforms nearly all state-of-the-art approaches in most metrics, and outperforms every tested approach on the ImageNet-VRD dataset.

The proposed approach in this paper is perhaps still naive, as all of the experts are tackling the same problem. In principle one could imagine dividing the problem into smaller subproblems (with certain experts only aimed at subject/object recognition, for example), to address multi-modal datasets or to explore hierarchical MoE architecture, in which a primary gating network chooses a combination of experts --- each of which itself is a secondary or tertiary MoE with its own respective set of experts and gating network \cite{shazeer2017}.

Finally, almost certainly the gating network itself would benefit from further scrutiny.  The gating network of this paper is the simplest possible choice, a single-layer feed-forward network, taking as input the {\em same} spatio-temporal object-tracklet features as are being provided to the experts. In a sense, it would seem that too much is being asked of the gating function, to go all the way from low-level input to expert-assessment output, such that the gating function tacitly must emulate or reproduce certain elements of expert behaviour. It would seem preferable to have the gating function operate at a higher / more abstracted level, and having certain aspects of expert-assessment made the responsibility of the expert networks themselves.

\newpage
\section{Acknowledgments}

We acknowledge the support of the Microsoft Office Media Group, advice from Chong Luo (Microsoft Research Asia), and the Alliance Program of the Natural Science \& Engineering Research Council of Canada.

\bibliographystyle{unsrtnat}
\bibliography{references}

\end{document}